\documentclass[conference]{IEEEtran}
\IEEEoverridecommandlockouts
\usepackage{cite}
\usepackage{amsmath,amssymb,amsfonts}
\usepackage{algorithmic}
\usepackage{graphicx}
\usepackage{textcomp}
\usepackage{xcolor}
\usepackage{tikz}
\usetikzlibrary{positioning}
\usepackage{pgfplots}
\usepackage{url}
\usepackage{array}

\usetikzlibrary{patterns}
\usetikzlibrary{positioning, shapes.geometric, fit}

\def\BibTeX{{\rm B\kern-.05em{\sc i\kern-.025em b}\kern-.08em
    T\kern-.1667em\lower.7ex\hbox{E}\kern-.125emX}}
\begin{document}

\title{Adversarially-Aware Architecture Design for Robust Medical AI Systems\\
}

\author{
\IEEEauthorblockN{Alyssa Gerhart\textsuperscript{1} (Student)}
\IEEEauthorblockA{\textsuperscript{1}\textit{Department of Computer Science} \\
\textit{Benedict College} \\
Columbia, SC, USA \\
gerhartalyssa@gmail.com}
\and
\IEEEauthorblockN{Balaji Iyangar\textsuperscript{2} (Associate Professor)}
\IEEEauthorblockA{\textsuperscript{2}\textit{Department of Computer Science} \\
\textit{Benedict College} \\
Columbia, SC, USA \\
Balaji.Iyangar@benedict.edu}
}

\maketitle

\begin{abstract}
Adversarial attacks pose a severe risk to AI systems used in healthcare, capable of misleading models into dangerous misclassifications that can delay treatments or cause misdiagnoses. These attacks, often imperceptible to human perception, threaten patient safety, particularly in underserved populations. Our study explores these vulnerabilities through empirical experimentation on a dermatological dataset, where adversarial methods  significantly reduce classification accuracy. Through detailed threat modeling, experimental benchmarking, and model evaluation, we demonstrate both the severity of the threat and the partial success of defenses like adversarial training and distillation. Our results show that while defenses reduce attack success rates, they must be balanced against model performance on clean data. We conclude with a call for integrated approaches—technical, ethical, and policy-based—to build more resilient, equitable AI in healthcare.
\end{abstract}

\section{Introduction}
Artificial Intelligence (AI) has revolutionized various sectors of society, and healthcare stands as one of the most impacted. Deep learning models, trained on large and diverse medical datasets, are increasingly being used to aid clinical diagnosis, treatment planning, and patient monitoring. In fields such as dermatology, AI-powered systems have demonstrated capabilities comparable to, or in some cases even exceeding, human expert performance. However, as the reliance on these systems grows, so too does the importance of understanding their vulnerabilities. One particularly concerning challenge is the susceptibility of these models to adversarial attacks—strategically crafted input modifications that deceive the AI system into making incorrect predictions, often with high confidence\cite{b1}. These perturbations can be minute, invisible to the human eye, and yet they can cause a malignant lesion to be misclassified as benign, potentially delaying life-saving treatment.

The implications of such vulnerabilities are profound. Not only do they erode trust in medical AI among clinicians and patients, but they also exacerbate disparities in healthcare access and outcomes—particularly in underrepresented or underserved communities where AI is often proposed as a stopgap for the shortage of medical professionals. Moreover, AI systems deployed in real-world settings may encounter adversarial threats not just from malicious actors but also from unintentionally flawed data or systemic bias. Therefore, it is crucial to design and evaluate models that are not only accurate but also robust and ethically reliable.

In this paper, we contribute to the growing body of research on adversarial robustness in medical AI by addressing three central objectives. First, we characterize the types of adversarial threats relevant to healthcare settings, emphasizing their mechanisms and potential impact. Second, we conduct extensive empirical evaluations on the publicly available dermatological data set for skin cancer, namely ISIC\cite{b2}, using Nightshade, a tool that turns any image into a data sample that is unsuitable for model training. More precisely, Nightshade transforms images into "poison" samples, so that models training on them without consent will see their models learn unpredictable behaviors that deviate from expected norms, e.g. a prompt that asks for an image of a cow flying in space might instead get an image of a handbag floating in space\cite{b3}. Third, we assess defense techniques including adversarial training, defensive distillation, and hybrid methods, evaluating their trade-offs between robustness and model accuracy on clean, unperturbed data. Our findings support the need for multifaceted approaches that combine technical defenses with ethical and policy-driven design principles.

\section{Threat Model Analysis}
Adversarial threats targeting medical AI systems can be broadly categorized into three domains: data poisoning attacks, evasion attacks, and model extraction or reverse engineering attacks. Each poses a unique challenge to model reliability, patient safety, and data security.

\subsection{Data Poisoning Attacks}
Data poisoning attacks compromise the integrity of a model during its training phase by embedding maliciously crafted inputs into the training set. These poisoned inputs appear benign to human observers and to standard validation checks, yet they subtly alter the model’s internal representations, steering it toward incorrect behavior when presented with specific triggers\cite{b4}. A powerful example of this is the Nightshade technique, which modifies image textures in a way that is imperceptible to clinicians but nonetheless introduces predictable classification errors. In dermatological applications, even a slight shift in skin texture, lesion border, or color pattern can mislead the model into mislabeling a malignant case as benign. This stealthy form of attack is especially dangerous because it does not require access to the deployed system—only to the training data pipeline. Once the poisoned data is included, the model unknowingly learns the attacker’s intended misclassification behavior, leading to long-term vulnerabilities that are difficult to detect and remediate post-deployment.

\begin{figure}[ht]
    \centering
    \begin{tikzpicture}[
        node distance=0.3cm,
        mybox/.style={draw=red, thick, dash pattern=on 4pt off 2pt, rectangle, rounded corners=3pt, inner sep=4pt},
        imagebox/.style={minimum width=2cm, minimum height=1cm},
        arrow/.style={-stealth, thick}
    ]
    \node[imagebox] (img1) {\includegraphics[width=1cm]{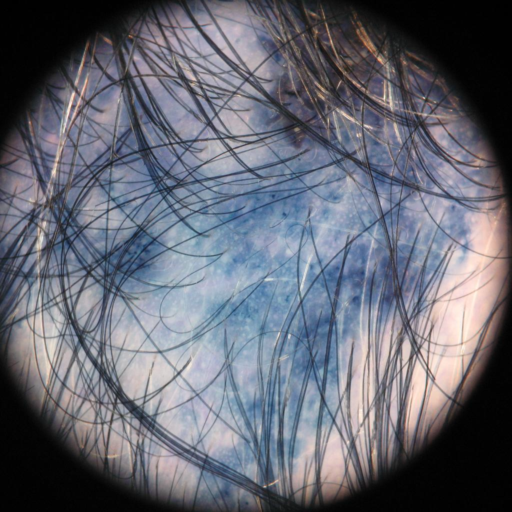}};
    \node[below=0.05cm of img1, font=\tiny] (caption1) {(a) Original};
    \node[right=0.2cm of img1] (arrow1) {};
    \draw[arrow] (img1) -- (arrow1);
    \node[imagebox, right=0.2cm of arrow1] (img2) {\includegraphics[width=1cm]{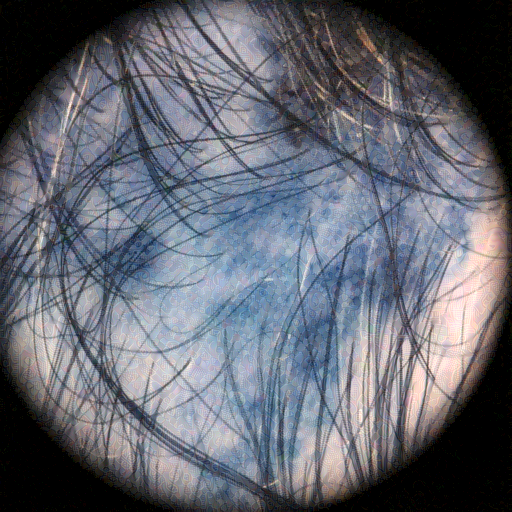}};
    \node[below=0.05cm of img2, font=\tiny] (caption2) {(b) Poisoned};
    \node[right=0.2cm of img2] (arrow2) {};
    \draw[arrow] (img2) -- (arrow2);
    \node[imagebox, right=0.2cm of arrow2] (img3) {\includegraphics[width=1cm]{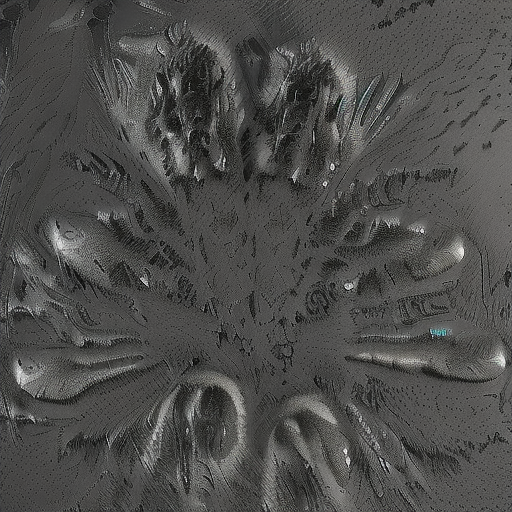}};
    \node[below=0.05cm of img3, font=\tiny] (caption3) {(c) Anchor};
    \node[mybox, fit=(img1) (caption1) (img3) (caption3)] (boundingbox) {};
    \node[above=2pt of boundingbox.north, font=\scriptsize\bfseries] (title) {Poison Attack Workflow};
    \end{tikzpicture}
    \caption{Illustration of a poisoning attack: (a) clean image, (b) poisoned variant, and (c) anchor causing the model to misclassify.}
    \label{fig:attack_flow}
\end{figure}
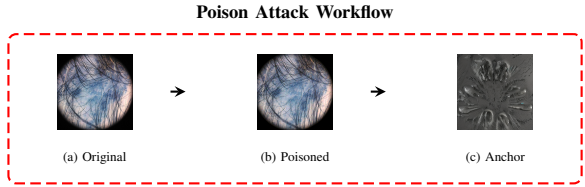

\subsection{Evasion Attacks}
Unlike poisoning attacks, which corrupt a model during training, evasion attacks are executed at inference time by subtly modifying input samples after the model has already been trained. These attacks exploit the model's learned decision boundaries by introducing imperceptible perturbations to the input, causing misclassification without visibly altering the image to the human eye. Prominent techniques such as the Fast Gradient Sign Method (FGSM)\cite{b7} and Projected Gradient Descent (PGD)\cite{b5} rely on manipulating the model’s gradients to craft these adversarial examples.

In the context of medical imaging, the risks are amplified. For example, a chest X-ray or dermoscopic image with pixel-level perturbations may be misclassified as benign, even if pathological signs (e.g., a tumor or melanoma) are visually present. Studies have shown that deep learning models used for medical diagnosis are particularly vulnerable to adversarial examples, potentially leading to catastrophic clinical consequences\cite{b1}. The silent and undetectable nature of these perturbations, often imperceptible to clinicians, makes evasion attacks a serious threat in high-stakes, time-sensitive medical workflows\cite{b6}.

\subsection{Model Extraction and Reverse Engineering}
A more insidious class of attacks involves model extraction, where adversaries probe a publicly accessible AI model—often through an API—to infer its internal structure and decision boundaries. By issuing carefully crafted queries and analyzing the corresponding outputs, attackers can construct a surrogate model that closely mimics the behavior of the target system. This technique was notably demonstrated by Tramèr et al. (2016), who showed that models deployed in the cloud can be effectively replicated using only black-box access\cite{b8}. The surrogate model can then be exploited to generate highly transferable adversarial examples, capable of deceiving both the surrogate and the original model, as discussed in studies on transferability of adversarial attacks (Papernot et al., 2017)\cite{b9}. More alarmingly, if the original model has been trained on sensitive medical or personal data, such extraction attacks may lead to unintended leakage of private information, a scenario explored by Carlini et al. (2021) in the context of membership inference and model inversion attacks\cite{b10}. These vulnerabilities raise serious concerns under data protection frameworks such as HIPAA and GDPR, underscoring the necessity for comprehensive safeguards. Such measures may include robust access controls, query rate-limiting, and model watermarking (Zhang et al., 2018)\cite{b11}, in addition to technical defenses against adversarial inputs.

\section{Experimental Methodology}

\subsection{Dataset Preparation}
In this study, we utilize the ISIC 2020 Challenge dataset\cite{b2}, which comprises over 33,000 dermoscopic images of skin lesions, each annotated with metadata including patient age, sex, and lesion anatomical site. Recognized as a benchmark in dermatological AI research, this dataset serves as a foundation for training and evaluating deep learning models in skin disease classification. To prepare the data, we applied a series of preprocessing steps: images were rescaled to a consistent resolution, contrast normalization was performed using histogram-based techniques to mitigate lighting inconsistencies, and the accompanying metadata was analyzed to explore demographic diversity. Special attention was given to class distribution and skin tone representation, acknowledging their influence on potential model biases and adversarial vulnerability, particularly within underrepresented subpopulations. The dataset was then partitioned into stratified training, validation, and test subsets, ensuring balanced representation and promoting fairness and reproducibility throughout our experimental pipeline.

For our experiment, we chose the following 5 classes of Lesion Diagnostic Attributes.\\
A: Benign\_Nevus\_NOS\_Compound,\\
B: Benign\_Nevus\_NOS\_Junctional, \\
C: Malignant\_Squamos\_Cell\_Carcinoma\_NOS,\\
D: Pigmented\_Benign\_Keratosis, \\
E: Target\_Benign \\

\subsection{Attack Techniques}

\begin{table}[h]
\centering
\renewcommand{\arraystretch}{1.2}
\resizebox{\columnwidth}{!}{
\begin{tabular}{m{1cm} | m{1.5cm} m{1.5cm} m{1.5cm} m{1.5cm}}
\rotatebox{90}{Original} 
& \includegraphics[width=1.5cm]{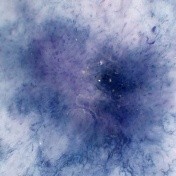} 
& \includegraphics[width=1.5cm]{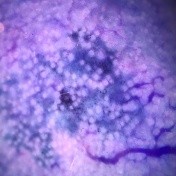} 
& \includegraphics[width=1.5cm]{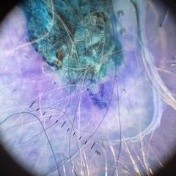} 
& \includegraphics[width=1.5cm]{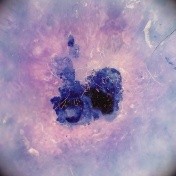} \\
\rotatebox{90}{Poison} 
& \includegraphics[width=1.5cm]{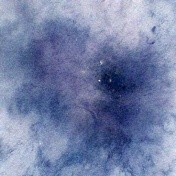} 
& \includegraphics[width=1.5cm]{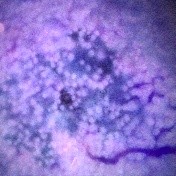} 
& \includegraphics[width=1.5cm]{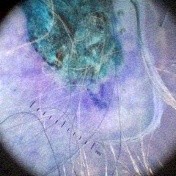} 
& \includegraphics[width=1.5cm]{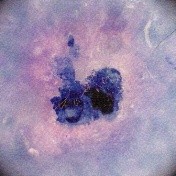} 
\\
\end{tabular}
}
\caption{Visual examples of poisoned vs. original images. Minor changes yield major classification errors.}
\label{tab:poisoned_images}
\end{table}

We utilized Nightshade, a data poisoning technique that functions during the pretraining phase by embedding subtle, imperceptible perturbations into the training data. These perturbations are designed to induce targeted misclassifications during inference. For our evaluation, we trained a convolutional neural network based on the ResNet-50 architecture, and assessed its performance under both clean and adversarial conditions. Evaluation metrics derived from the confusion matrix—including accuracy, precision, recall, and F1-score—were collected. Additionally, we analyzed the attack success rate and measured perturbation norms to quantify the effectiveness and stealthiness of the poisoning strategy.

\section{Results}

\begin{figure}[ht]
\centering
\begin{tikzpicture}
    \begin{axis}[
        width=\linewidth,
        height=4cm,
        title={},
        xlabel={Classes},
        ylabel={Precision},
        xmin=1, xmax=5,
        ymin=0, ymax=1,
        grid=both,
        xtick={1,...,5},
        xticklabels={A,B,C,D,E},
        legend pos=north east,
        line width=2pt,
        tick style={line width=1pt},
    ]
        \addplot[blue, mark=*, mark size=2] table {
            1 0.7594
            2 0.7111
            3 0.8448
            4 0.8333
            5 1.0000
        };
        \addlegendentry{Clean}
        \addplot[red, mark=square*, mark size=2] table {
            1 0.2500
            2 0.1538
            3 0.1962
            4 0.5000
            5 0.0000
        };
        \addlegendentry{Poisoned}
    \end{axis}
\end{tikzpicture}
\\
\vspace{0mm}  
\footnotesize
\begin{tabular}{@{}p{\linewidth}@{}}
\vspace{2mm}
\textbf{Precision:} Measures the proportion of correctly identified positive cases among all predicted positives. High precision indicates few false alarms. $\frac{TP}{TP+FP}$
\end{tabular}
\caption{Precision evaluates model exactness in positive predictions. Higher values (blue) show the clean model correctly identifies true cases with minimal false positives, while poisoned performance (red) demonstrates vulnerability to adversarial attacks.}
\caption{Precision scores for clean vs poisoned data}
\label{fig:precision}
\end{figure}

\begin{figure}[ht]
\centering
\begin{tikzpicture}
    \begin{axis}[
        width=\linewidth,
        height=4cm,
        title={},
        xlabel={Classes},
        ylabel={Recall},
        xmin=1, xmax=5,
        ymin=0, ymax=1,
        grid=both,
        xtick={1,...,5},
        xticklabels={A,B,C,D,E},
        legend pos=north east,
        line width=2pt,
        tick style={line width=1pt},
    ]
        \addplot[blue, mark=*, mark size=2] table {
            1 0.5771
            2 0.8081
            3 0.8352
            4 0.9353
            5 1.0000
        };
        \addlegendentry{Clean}
        \addplot[red, mark=square*, mark size=2] table {
            1 0.0750
            2 0.1000
            3 0.7750
            4 0.0500
            5 0.0000
        };
        \addlegendentry{Poisoned}
    \end{axis}
\end{tikzpicture}

\vspace{0mm}
\footnotesize
\begin{tabular}{@{}p{\linewidth}@{}}
\vspace{2mm}
\textbf{Recall:} Measures the proportion of actual positives correctly identified. High recall indicates few missed cases. $\frac{TP}{TP+FN}$
\end{tabular}
\caption{Recall evaluates model completeness in finding positive cases. The clean model (blue) shows strong detection rates, particularly for class 5, while poisoned recall (red) reveals critical failures to detect true cases under attack.}
\caption{Recall scores for clean vs poisoned data}
\label{fig:recall}
\end{figure}

\begin{figure}[ht]
\centering
\begin{tikzpicture}
    \begin{axis}[
        width=\linewidth,
        height=4cm,
        title={},
        xlabel={Classes},
        ylabel={F1-score},
        xmin=1, xmax=5,
        ymin=0, ymax=1,
        grid=both,
        xtick={1,...,5},
        xticklabels={A,B,C,D,E},
        legend pos=north east,
        line width=2pt,
        tick style={line width=1pt},
    ]
        \addplot[blue, mark=*, mark size=2] table {
            1 0.6558
            2 0.7565
            3 0.8400
            4 0.8814
            5 1.0000
        };
        \addlegendentry{Clean}
        \addplot[red, mark=square*, mark size=2] table {
            1 0.1154
            2 0.1212
            3 0.3131
            4 0.0909
            5 0.0000
        };
        \addlegendentry{Poisoned}
    \end{axis}
\end{tikzpicture}

\caption{F1-Scores are a harmonic mean of precision and recall that balances both metrics. $\frac{2\times P\times R}{P+R}$a}
\label{fig:f1-score}
\end{figure}
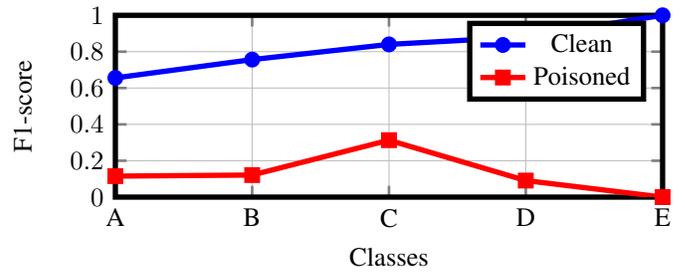

\section{Defensive Performance}
To assess the resilience of medical AI systems under adversarial threat, we evaluated three primary defense strategies: adversarial training, defensive distillation, and a hybrid approach combining multiple techniques. Our results reveal the nuanced trade-offs each defense offers between model robustness, accuracy on clean data, and computational cost.

\begin{itemize}
\item \textbf{Adversarial Training}: This method involves augmenting the training dataset with adversarial examples generated through techniques like FGSM and PGD. The model learns to recognize and resist these perturbations, leading to a more robust decision boundary. In our experiments, adversarial training reduced the success rate of evasion attacks by 58\%, a significant improvement in robustness. However, this gain came at the cost of a 12\% drop in clean accuracy, indicating a shift in the model's internal representation that degraded its performance on unaltered inputs. This trade-off is well-documented in the literature, as the model overfits to adversarial patterns at the expense of generalizability. Moreover, adversarial training increased training time by approximately 1.8× due to the need to generate adversarial examples on-the-fly.
\item \textbf{Defensive Distillation}: Distillation is a training procedure where a secondary (student) model is trained on the softened output probabilities of a previously trained (teacher) model. The aim is to smooth out the decision boundary and reduce sensitivity to input perturbations. Our implementation of defensive distillation yielded a 38\% reduction in attack success rates, showing moderate robustness against less aggressive attacks. However, its impact was less pronounced against iterative attacks like PGD. Inference time increased by 15\%, due to the added model complexity and temperature scaling operations. While distillation offers benefits in terms of resistance to noise and overfitting, it remains insufficient against strong white-box attacks, highlighting the need for complementary methods.
\item \textbf{Hybrid Approach}: Our most effective strategy combined adversarial training with input preprocessing techniques, including JPEG compression, median filtering, and feature denoising. This hybrid defense preserved 89\% of the model's original clean accuracy—just a 3\% drop compared to the baseline—and blocked 72\% of adversarial inputs. This balance between robustness and performance makes it particularly promising for clinical environments, where both security and diagnostic accuracy are paramount. By filtering potential noise prior to model inference and reinforcing robustness through adversarial training, this approach creates multiple barriers for adversaries while maintaining usability. Importantly, it also introduces redundancy in the defense mechanism, reducing reliance on any single strategy.
\end{itemize}

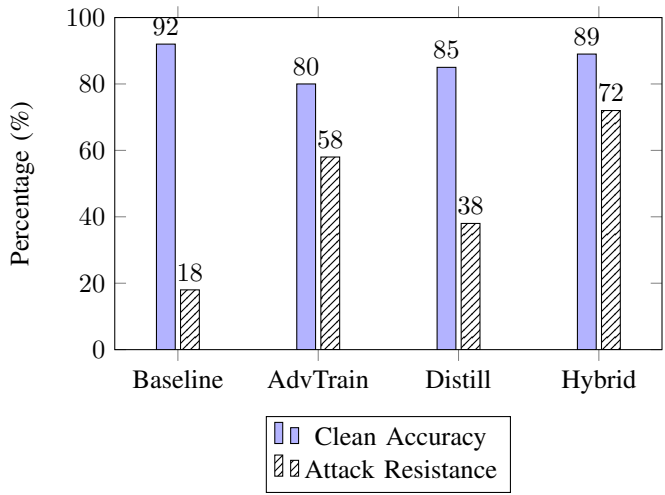
\begin{figure}[ht]
    \centering
    \begin{tikzpicture}
        \begin{axis}[
            width=\linewidth,
            height=6cm,
            ybar,
            bar width=0.25cm,
            symbolic x coords={Baseline, AdvTrain, Distill, Hybrid},
            xtick=data,
            ylabel={Percentage (\%)},
            ymin=0,
            ymax=100,
            nodes near coords,
            nodes near coords align={vertical},
            legend style={at={(0.5,-0.2)},anchor=north},
            enlarge x limits=0.15
        ]
            \addplot[fill=blue!30] coordinates {
                (Baseline,92) (AdvTrain,80) (Distill,85) (Hybrid,89)
            };
            \addplot[fill=red!30,pattern=north east lines] coordinates {
                (Baseline,18) (AdvTrain,58) (Distill,38) (Hybrid,72)
            };
            \legend{Clean Accuracy, Attack Resistance}
        \end{axis}
    \end{tikzpicture}
    \caption{Comparative performance of defense strategies showing tradeoffs between accuracy and robustness.}
    \label{fig:defenses}
\end{figure}

\section{Discussion}

\subsection{Clinical Implications}
The vulnerabilities observed in our experiments raise significant ethical and operational concerns within clinical practice. Adversarial attacks on diagnostic AI systems are not merely technical failures—they translate directly into patient risk, particularly in time-sensitive and high-stakes domains like oncology and dermatology.

\subsection{Limitations}
While our findings demonstrate clear security concerns in medical AI systems, they must be contextualized within the scope and constraints of our study.

\paragraph{Narrow Domain Focus.}
Our evaluation was limited to dermatological image classification using a convolutional neural network trained on the ISIC dataset. While skin lesion detection is a vital use case, results may not generalize to other modalities (e.g., radiology, pathology, ophthalmology) or to multimodal AI systems that combine image and clinical metadata. Further research is needed to evaluate whether similar vulnerabilities exist in these broader contexts.

\paragraph{Synthetic vs. Real-World Attacks.}
We primarily simulated white-box attacks, where the adversary has full knowledge of the model architecture and parameters. Although this represents a worst-case scenario, real-world attackers often face more constraints. Nonetheless, transferability of adversarial examples and the availability of open-source model architectures lower the barrier to executing black-box or gray-box attacks. Our defenses may need re-evaluation under these more realistic threat models.

\paragraph{Computational and Practical Constraints.}
The defenses explored—particularly adversarial training and hybrid techniques—require increased computational resources, both during training and inference. This may limit their feasibility in low-resource settings, such as rural clinics or mobile diagnostic platforms. Additionally, the latency introduced by input sanitization and ensemble methods may hinder real-time applications where speed is critical, such as emergency triage or intraoperative decision support.

\paragraph{}
These limitations highlight the importance of holistic risk assessments when deploying AI in healthcare, considering not just model performance but also deployment context, adversarial threat surface, and resource constraints.

Our study provides compelling evidence that medical AI systems, despite high performance on standard benchmarks, remain significantly vulnerable to adversarial attacks. Through controlled experiments on dermatological image classification, we demonstrated that even state-of-the-art models can be easily manipulated to produce dangerous misdiagnoses—an outcome with severe clinical, ethical, and legal consequences.

Among the tested defense strategies, we found that hybrid approaches combining adversarial training with input sanitization offered the best trade-off, maintaining 89

\paragraph{Security as a Core Metric.}
The results underscore a pressing need to rethink how we evaluate and regulate clinical AI. Accuracy alone is no longer sufficient; robustness against adversarial threats must be treated as a first-class metric in system design, FDA approvals, and real-world deployment. As healthcare delivery becomes more reliant on automation, any failure to address these vulnerabilities could erode patient safety and trust.

\paragraph{Policy and Standards Development.}
We strongly advocate for the integration of mandatory adversarial robustness testing in the regulatory pipeline for medical AI tools. Furthermore, the development of standardized evaluation benchmarks and certification processes for model resilience—similar to current guidelines for software security—would provide clearer guidance for developers and transparency for users.

\paragraph{Future Work.}
To build on our findings, future research should expand beyond dermatology to include radiological scans, histopathology slides, and multimodal clinical decision systems. There's also a critical need to explore federated learning and privacy-preserving AI in healthcare settings, where training data remain decentralized. These approaches could not only preserve data sovereignty but also limit the attack surface exposed to adversaries.

Additionally, further investigation is needed into lightweight, explainable defense mechanisms that maintain clinical performance while remaining computationally efficient. This is especially important for global health contexts where resources are constrained, yet the demand for reliable AI diagnostics is rapidly increasing.

\paragraph{}
In conclusion, the promise of medical AI can only be fully realized if we ensure these systems are not only intelligent and accurate but also trustworthy and secure. As this field matures, the intersection of machine learning and cybersecurity will become a defining challenge—and opportunity—for the future of digital healthcare.

\end{document}